\documentclass{article} 
\usepackage{firas2026_conference,times}

\usepackage{amsmath,amsfonts,bm}

\def\eqref#1{equation~\ref{#1}}

\def\1{\bm{1}}

\DeclareMathAlphabet{\mathsfit}{\encodingdefault}{\sfdefault}{m}{sl}
\SetMathAlphabet{\mathsfit}{bold}{\encodingdefault}{\sfdefault}{bx}{n}

\usepackage{hyperref}
\usepackage{url}
\usepackage{svg}
\svgsetup{inkscapelatex=false}
\usepackage{graphicx}
\usepackage{booktabs}
\usepackage{array}
\usepackage{caption}
\usepackage{float}

\title{Motion Capture is Not the Target Domain: Scaling Synthetic Data for Learning Motion Representations}

\author{Firas Darwish\thanks{Corresponding Author} \\
Department of Statistics\\
University of Oxford\\
Oxford, UK \\
\texttt{\{firas.darwish\}@worc.ox.ac.uk} \\
\And
George Nicholson \\
Department of Statistics \\
Big Data Institute \\
University of Oxford \\
Oxford, UK \\
\texttt{\{george.nicholson\}@stats.ox.ac.uk}
\AND
Aiden Doherty\thanks{Joint supervision.} \\
Nuffield Department of Population Health \\
Big Data Institute \\
University of Oxford \\
Oxford, UK  \\
\texttt{\{aiden.doherty\}@ndph.ox.ac.uk}
\And
Hang Yuan\footnotemark[2] \\
Nuffield Department of Population Health \\
Big Data Institute \\
University of Oxford \\
Oxford, UK  \\
\texttt{\{hang.yuan\}@ndph.ox.ac.uk}
}

\firasfinalcopy 
\begin{document}

\maketitle

\begin{abstract}
Synthetic data offers a compelling path to scalable pretraining when real-world data is scarce, but models pretrained on synthetic data often fail to transfer reliably to deployment settings. We study this problem in full-body human motion, where large-scale data collection is infeasible but essential for wearable-based Human Activity Recognition (HAR), and where synthetic motion can be generated from motion-capture–derived representations. We pretrain motion time-series models using such synthetic data and evaluate their transfer across diverse downstream HAR tasks. Our results show that synthetic pretraining improves generalisation when mixed with real data or scaled sufficiently. We also demonstrate that large-scale motion-capture pretraining yields only marginal gains due to domain mismatch with wearable signals, clarifying key sim-to-real challenges and the limits and opportunities of synthetic motion data for transferable HAR representations.
\end{abstract}

\section{Introduction}
\label{ref:Introduction}

Synthetic data holds significant promise for addressing data scarcity in machine learning, particularly in regimes where collecting large-scale, diverse real-world datasets is expensive, slow, or impractical \citep{ONETO2025130521, BRODA2025113728, lu2023machine}. By complementing (or in some cases substituting) real data, synthetic data enables scalable training across tasks and environments while offering greater control over data diversity and coverage.

Recent progress across domains highlights the utility of synthetic data, with successful applications in robotics \citep{simtorealsurvey, cheng2025generalizable}, autonomous driving \citep{Song_2024, hu2023simulationhelpsautonomousdrivinga}, and Large Language Model (LLM) pretraining \citep{kang2025demystifyingsyntheticdatallm, alcoba-inciarte-etal-2024-utility}. Despite these successes, models trained primarily on synthetic data often struggle to transfer reliably to real-world settings due to the sim-to-real gap, where mismatches in perceptual or dynamical details lead to degraded performance at deployment \citep{hu2023simulationhelpsautonomousdrivinga, simtorealsurvey}.

Human motion (particularly full-body human motion) is another domain where scaling data collection is essential. Human Activity Recognition (HAR) models use motion data from wearable devices to classify activities over time, but models trained and evaluated on the same datasets often generalise poorly due to systematic variations in sensors, subjects, and environments \citep{chen2021deeplearningsensorbasedhuman}. Consequently, pretraining on large and diverse motion data is critical for learning representations that transfer effectively across HAR tasks.

However, collecting human motion data at scale remains infeasible in practice \citep{lin2024motionxlargescale3dexpressive}. Motion capture datasets, while accurate, are limited in both diversity and volume and are largely restricted to controlled laboratory environments \citep{wang20143d}, while large-scale wearable data collection in the wild is limited to wrist motion \citep{ma2024nymeriamassivecollectionmultimodal}.

At the same time, human motion is a natural candidate for synthetic generation via simulation or generative modelling because it is governed by well-defined physical and kinematic constraints that must be maintained over extended temporal horizons, such as balance, contact dynamics, and joint limits \citep{yuan2023physdiffphysicsguidedhumanmotion}.

In this paper, we investigate the utility of using synthetic data in pretraining motion time-series models that are used downstream in different HAR tasks. Our core contributions in this paper can be summarised as:
\begin{enumerate}
    \item We offer a framework to apply the sim-to-real paradigm in the context of wearables, and formulate the core challenges that arise when representing human motion.
    \item We demonstrate that pretraining motion time-series models on synthetic data improves generalisation to downstream HAR tasks; however, we do not observe the same scaling laws compared to training on real data \citep{narayanswamy2024scalingwearablefoundationmodels}.
    \item We show that scaling motion-capture-based pretraining leads to marginal gains in HAR tasks, exhibiting evidence of a domain mismatch with the downstream wearable sensor signals that limits the transferability of motion-capture representations.
\end{enumerate}

\section{Related Work}
\label{ref:RelatedWork}

\subsection{Pretraining Motion Time-Series Models}

Pretraining motion time-series models has gained momentum as a means of learning transferable representations for HAR, typically through self-supervised objectives such as masked modelling \citep{LIU2025121576, SSL_interpolation_forecasting_mask}, contrastive learning \citep{zhang2024unimtsunifiedpretrainingmotion, xu2025relconrelativecontrastivelearning}, multimodal alignment \citep{girdhar2023imagebindembeddingspacebind}, or transformation discrimination \citep{yuan2024self}. Despite this progress, generalisation across HAR datasets remains challenging due to systematic variations in subjects, activities, and sensing setups: physiological and biomechanical differences between wearers alter the mapping from latent activity to sensor signals \citep{Jimale2021SubjectVI, CalatravaNicols2025EmbeddedIV}, activity sets vary across datasets with only partial overlap, and device heterogeneity introduces additional distribution shifts \citep{sensor_placement_variation, Stisen2015SmartDA}. 

In this work, we adopt the UniMTS pretraining and evaluation framework \citep{zhang2024unimtsunifiedpretrainingmotion} because its contrastive formulation achieves strong performance while explicitly addressing these sources of variation through full-body Inertial Measurement Unit (IMU) pretraining, random joint masking, and rotation-based augmentation for device invariance; moreover, its text–motion alignment enables compatibility with text-conditioned motion generation, and its evaluation across 18 diverse HAR datasets provides a rigorous test of generalisation.

\subsection{Human Motion Generation}

Human motion generation has been extensively studied across a wide range of generative paradigms, conditioning modalities, and evaluation protocols \citep{Zhu2023HumanMG, Khani2025MotionGA}. Prior work explores diverse modelling approaches, including GANs \citep{GANs1, styleGAN}, VAEs \citep{VAE1, vae2, T2M}, diffusion models \citep{MotionDiffuse, MDM}, and autoregressive architectures \citep{wang2025motiondreameronetomanymotionsynthesis, pinyoanuntapong2024mmmgenerativemaskedmotion}, with generation conditioned on text \citep{MDM, MotionDiffuse, T2M}, discrete action labels \citep{A2M}, audio signals such as music or speech \citep{Qi_2023, speech_gesture_generation}, or scene representations to ensure physical and environmental consistency \citep{Marcard2018RecoveringA3, Zhang2021EgoBodyHB}. 

Evaluation typically combines qualitative human evaluations \citep{Andreou2024LEADLR} with quantitative fidelity metrics that measure similarity to ground-truth motion \citep{Raab2023SingleMD}, assessments of physical plausibility (like ground contact, penetration, and foot sliding) \citep{Li2023AAMDMAA}, measures of motion diversity across and within activities \citep{Barquero2022BeLFusionLD}, and alignment with conditioning inputs \citep{Li2024LaMPLP}. As a result, contemporary motion generators produce data of sufficient fidelity and coverage to serve as a viable source of human motion data.

\subsection{Synthetic Data in Wearables}


Prior work has explored synthetic data generation for wearable-based HAR using a range of generative approaches. GAN-based methods have been shown to effectively substitute or augment real data for gesture and activity recognition, improving performance in low-data and class-imbalanced settings \citep{kontogiannis2024exploring, li2020activitygan, alharbi2020synthetic, de2023exploring}. More recent work demonstrates that synthetic multivariate time-series generated from wearable sensors can improve HAR accuracy in data-scarce regimes and across larger activity sets \citep{silva2025synthetic, lupion2024data}, while augmentation with in-the-wild smartwatch data further improves performance \citep{desmet2024hydra}.

However, these works focus exclusively on supervised learning settings, and do not examine the role of synthetic data in pretraining motion time-series models or its impact on representation learning and transfer across HAR tasks.

\section{Problem Formulation}
\label{ref:ProblemFormulation}

\subsection{Motion Representation}
\label{ref:MotionRepresentation}

\begin{figure}[t]
  \centering
  \includegraphics[
    width=\linewidth,
  ]{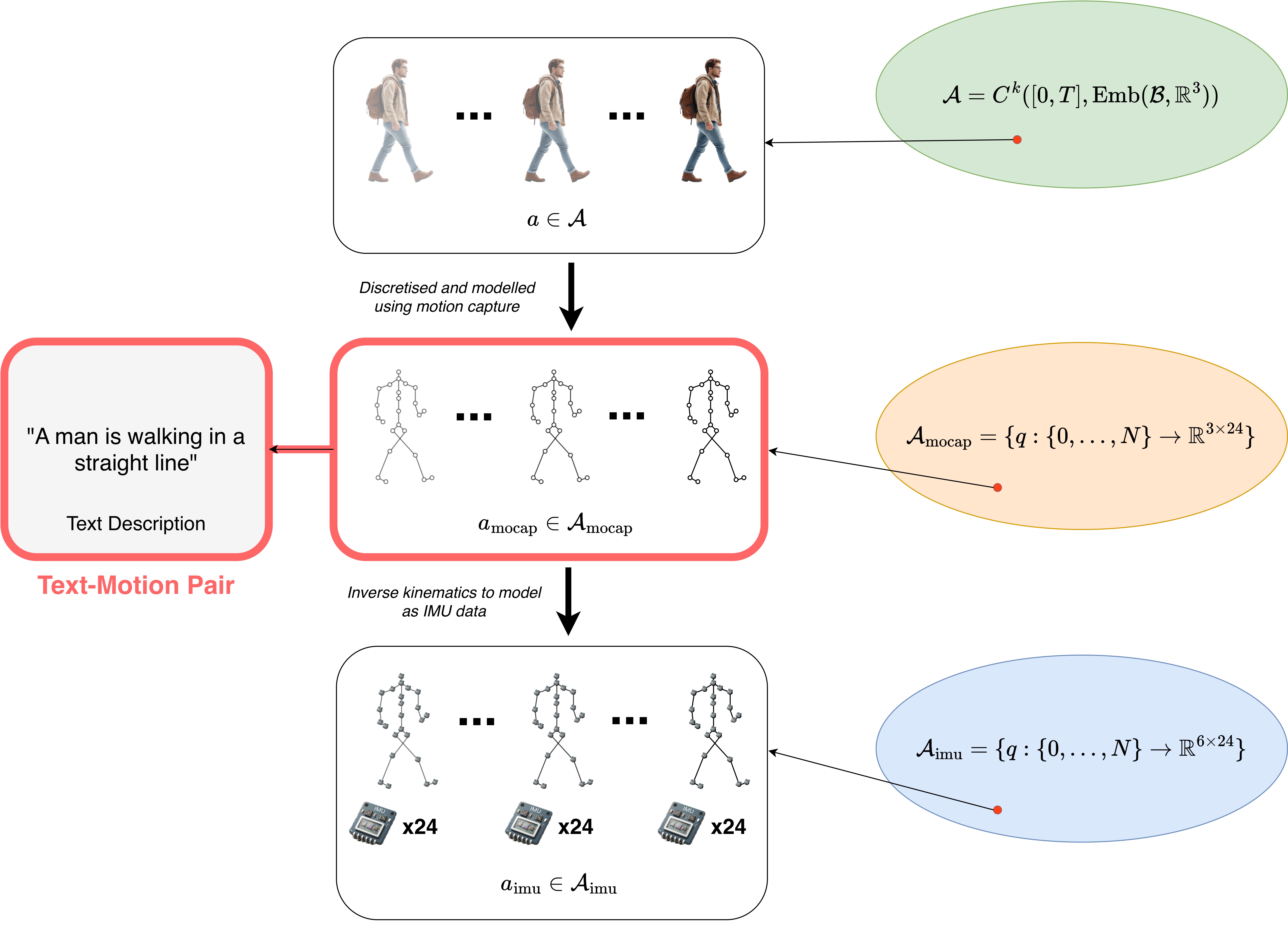}
  \caption{\textbf{Modelling Human Motion.} Human motion modelling relies on selecting a number of joints to track over a period of time, forming a discretised representation of human motion. This motion can be labelled and annotated with text. Using inverse kinematics, one can simulate accelerometer and gyroscope readings for each of those joints over time that can be used for wearables pretraining.}
  \label{fig:ProblemFormulationImage}
\end{figure}

We model human actions as continuous motion of a deformable body in physical space. Let $\mathcal{B}$ denote the human body, represented as a compact three-dimensional material manifold whose points correspond to material particles of the body. At any time $t \in [0,T]$, a body configuration is given by a smooth embedding $$ q(t) : \mathcal{B} \rightarrow \mathbb{R}^3, $$ which assigns a spatial position to every body point.

An action is then defined as a time-indexed family of such configurations (a curve in configuration space). Formally, an action is a function $$ a \in C^k([0,T],\ \text{Emb}(\mathcal{B},\mathbb{R}^3)), $$ where $C^k$ denotes $k$-times continuously differentiable functions in time, and $\text{Emb}(\mathcal{B},\mathbb{R}^3)$ is the space of smooth embeddings of the body into three-dimensional physical space. The regularity condition $k \geq 2$ ensures that velocity and acceleration fields exist and vary continuously for every body point.

The space of all human actions is therefore the infinite-dimensional function space $$ \mathcal{A} = C^k([0,T],\ \text{Emb}(\mathcal{B},\mathbb{R}^3)). $$ While the action space $\mathcal{A}$ provides a holistic representation of human motion,  real-world sensing and learning systems such as motion capture operate on finite, discretised observations that partially captures this action space.

Let $N \in \mathbb{N}$ denote the number of discrete time steps and let a fixed set of 24 anatomical landmarks (joints) be selected on the body, following the SMPL representation~\citep{SMPL, zhang2024unimtsunifiedpretrainingmotion, T2M}. At each time step, the spatial configuration of the body is represented by the three-dimensional positions of these joints. A discretised action is then defined as a mapping $$ q : \{0, \dots, N\} \rightarrow \mathbb{R}^{3 \times 24}, $$ where $q(n)$ specifies the stacked $x$, $y$, $z$ coordinates of the 24 tracked joints at discrete time index $n$.

While $\mathcal{A}_{\text{mocap}}$ provides a discretised representation of human motion in terms of joint positions, many downstream HAR systems operate on data collected from wearable inertial sensors. To bridge this gap, we introduce an IMU-based action space that aligns with the measurement modality of such sensors.

Specifically, we define the IMU action space as $$\mathcal{A}_{\text{imu}} = \{q:\{0,\dots,N\} \rightarrow \mathbb{R}^{6 \times 24}\},$$ where each of the 24 joints is associated with a six-dimensional inertial measurement vector consisting of tri-axial linear acceleration and tri-axial angular velocity. For each discrete time index $n$, $q(n)$ represents the stacked inertial measurements across all joints.

The mapping from $\mathcal{A}_{\text{mocap}}$ to $\mathcal{A}_{\text{imu}}$ is obtained via inverse kinematics and differentiation, converting joint position trajectories into joint-centric accelerometer and gyroscope signals \citep{zhang2024unimtsunifiedpretrainingmotion}. This transformation preserves the underlying motion while expressing it in a sensor-centric coordinate system consistent with IMUs.

\subsection{Text-to-Motion Generation}
\label{ref:TextToMotionGeneration}

We consider a dataset of paired natural language descriptions and discretised motion capture sequences, which can be used to train a text-conditioned generative model that learns a distribution over motions in the motion capture action space $\mathcal{A}_{\text{mocap}}$ \citep{T2M}. Because this mapping is inherently non-deterministic, a single text description can correspond to multiple valid motion realisations, and sampling from the learned distribution produces diverse motion trajectories consistent with the same prompt \citep{petrovich2022temosgeneratingdiversehuman, zhang2024unimtsunifiedpretrainingmotion}.

Given a trained text-conditioned motion model and a predefined set of text prompts, we generate a synthetic motion capture dataset by sampling $K$ motion trajectories per prompt. The resulting motion sequences are then converted into simulated IMU signals using the same inverse kinematics procedure described earlier, expressing each sample in the IMU action space $\mathcal{A}_{\text{imu}}$, which can be used to pretrain wearable motion time-series models \citep{zhang2024unimtsunifiedpretrainingmotion}.

\subsection{Pretraining Motion Time-Series Models}
\label{ref:PretrainingMotionTSModels}

\subsubsection{Pretraining}
\label{subsubsec:Pretraining}

To learn motion representations that generalise across downstream HAR tasks, we pretrain a motion encoder using contrastive learning on paired text–motion data, following \citet{zhang2024unimtsunifiedpretrainingmotion}. A dual-encoder architecture maps text descriptions and their corresponding IMU motion sequences (drawn from either real datasets or synthetically generated pairs) into a shared embedding space, and training optimises a contrastive objective that increases similarity between matched text–motion pairs while reducing similarity for mismatched pairs within a batch.

\subsubsection{Evaluation}
\label{subsubsec:Evaluation}

Given multiple HAR datasets containing labelled motion sequences in the IMU action space $\mathcal{A}_{\text{imu}}$ and a pretrained motion encoder, we evaluate downstream performance under both 0-shot and $k$-shot settings following UniMTS \citep{zhang2024unimtsunifiedpretrainingmotion}. Because HAR datasets vary in sensor placement and number, we align each dataset with the pretrained 24-joint motion encoder by retaining channels corresponding to observed joints and zeroing out unobserved ones.

In the 0-shot setting, the motion encoder is frozen and classification is performed by matching motion embeddings to text embeddings of activity labels via similarity scores. In the $k$-shot setting, we fine-tune the pretrained motion encoder with a linear classification head using $k$ labelled examples per activity class from the target dataset, and then evaluate performance on the dataset’s test split.

\section{Experiments}
\label{ref:Experiments}

\subsection{Implementation Details}
\label{subsection:ImplementationDetails}

We generate our synthetic motion data using state of the art text-to-motion models that were all trained on labelled motion capture sequences from HumanML3D \citep{T2M}: (1) Human Motion Diffusion Model (MDM) \citep{MDM}, (2) Text-to-motion (T2M) \citep{T2M}, and (3) MotionDiffuse \citep{MotionDiffuse}.

Our real-world data consists of 24,661 motion sequences extracted from HumanML3D motion capture dataset and transformed into $A_{\text{imu}}$ \citep{T2M}. Each of these motion sequences is paired with a text file containing around 6 sentences, with each sentence providing an individual annotated description of the motion \citep{zhang2024unimtsunifiedpretrainingmotion}. Using each text-to-motion model, we randomly sample one description from the associated text file and generate a motion conditioned on that sentence. This procedure is applied to all 24,661 text files, yielding 24,661 generated motion sequences per pass. We repeat this process multiple times to produce synthetic datasets at increasing multiples of the real-world data volume. The resulting synthetic dataset sizes can be found in Table~\ref{tab:data_volume} of Appendix~\ref{sec:Appendix}.

Following from \cite{zhang2024unimtsunifiedpretrainingmotion}, our motion encoder is implemented as a 4.94M parameter spatio-temporal graph convolutional network that processes joint-level accelerometry signals over a fixed temporal window. We pretrain the model for 100 epochs using a batch size of 64, learning rate of $10^{-4}$, and temperature parameter of 0.1. We also apply data augmentation in the form of random rotations of the accelerometry signals and random masking of a subset of joints during each training iteration. For the text modality, we use the pretrained CLIP ViT-B/32 text encoder \citep{radford2021learningtransferablevisualmodels}, which is fine-tuned jointly during pretraining.

To evaluate the pretrained models, we compute the macro-averaged F1 score within each dataset, and then report the mean of these scores across all 18 HAR datasets from \citet{zhang2024unimtsunifiedpretrainingmotion} for the 0-shot, 1-shot, 2-shot, 3-shot, 5-shot, and 10-shot settings.

All synthetic data generation, pretraining, and evaluation experiments are conducted using approximately 12 NVIDIA A100 GPUs, for a total of 60 GPU-days.

\subsection{Results}
\label{subsection:Results}

\subsubsection{Pretraining with Mixed Data}
\label{subsubsection:PretrainingwithMixedData}

We study the effect of pretraining data composition by comparing real-world data, synthetic data, and their mixtures while holding the total pretraining volume fixed, allowing us to evaluate both the standalone utility of synthetic motions and their ability to complement real-world data.

Across all settings, we pretrain on a fixed total of 24,660 text–motion pairs, with the following configurations:
\begin{itemize}
\item \textbf{Real-World Baseline:} 24,660 real-world text–motion pairs, matching the setup in \cite{zhang2024unimtsunifiedpretrainingmotion}.
\item \textbf{Purely Synthetic Data:} 24,660 synthetic text–motion pairs generated by each text-to-motion model (Table~\ref{tab:data_volume}).
\item \textbf{Mixed Data:} A randomly sampled 50/50 mixture of real-world and synthetic text–motion pairs.
\end{itemize}
Results are reported in Table~\ref{tab:data_mixing_results}. The overall gains from mixed pretraining relative to the real-world baseline align with what \cite{kang2025demystifyingsyntheticdatallm} demonstrated in LLM pretraining. Beyond that, models pretrained on purely synthetic data achieve performance close to the real-world baseline. Both synthetic-only and mixed pretraining perform particularly well in higher-shot regimes (3-, 5-, and 10-shot). Across synthetic pretraining settings, we further observe that no single text-to-motion generative model consistently outperforms the others across all evaluation regimes. Performance differences between synthetic datasets are relatively small, indicating that different generators produce motion data of comparable utility for motion time-series pretraining.

\begin{table}[t]
\centering
\caption{\textbf{Synthetic data effectively complements real motion data during pretraining.} Across downstream HAR tasks, pretraining on mixed real and synthetic data tends to improve generalisation relative to real-world data alone, and improvements are most evident in higher-shot regimes. Reported scores are mean macro-averaged F1 across datasets; \textbf{bold} and \underline{underlined} values denote best and second-best results per column. All results are obtained using a fixed random seed across runs.}
\label{tab:data_mixing_results}
\addtolength{\tabcolsep}{4pt} 
\begin{tabular}{l ccccccc}
\toprule
\textbf{Method} & \textbf{0-shot} & \textbf{1-shot} & \textbf{2-shot} & \textbf{3-shot} & \textbf{5-shot} & \textbf{10-shot} \\  \midrule

\textbf{Real-World Baseline} & \textbf{0.3070} & \underline{0.5445} & \textbf{0.6243} & 0.6611 & 0.7041 & 0.7667 \\ \midrule

\textbf{Purely Synthetic Data} \\
\quad MDM & 0.2460 & 0.5120 & \underline{0.6160} & 0.6559 & 0.6974 & 0.7653 \\
\quad T2M & 0.2606 & 0.5229 & 0.6100 & 0.6551 & 0.7024 & 0.7649 \\
\quad MotionDiffuse & 0.2738 & 0.5363 & 0.6060 & 0.6548 & \textbf{0.7053} & \underline{0.7739} \\ \midrule

\textbf{Mixed Data} \\
\quad MDM & \underline{0.3002} & 0.5337 & 0.6014 & \underline{0.6738} & \underline{0.7049} & 0.7702 \\
\quad T2M & 0.2901 & \textbf{0.5539} & 0.6139 & \textbf{0.6931} & 0.6970 & 0.7687 \\
\quad MotionDiffuse & 0.2732 & 0.5342 & 0.6099 & 0.6707 & 0.7004 & \textbf{0.7931} \\ 
\bottomrule
\end{tabular}
\end{table}

\subsubsection{Pretraining with Synthetic Data}
\label{subsubsection:PretrainingwithSyntheticData}

\begin{table}[t]
\centering
\caption{\textbf{Scaling synthetic data improves downstream HAR performance.} At sufficient scale ($8\times$), synthetic pretraining outperforms real-world pretraining in all fine-tuned settings, with the exception of the 0-shot regime. Reported scores are mean macro-averaged F1 across datasets; \textbf{bold} and \underline{underlined} values denote best and second-best results per column, respectively. All results are obtained using a fixed random seed across runs.}
\label{tab:synthetic_scaling_results}
\addtolength{\tabcolsep}{3pt} 
\begin{tabular}{l cccccc}
\toprule
\textbf{Method} & \textbf{0-shot} & \textbf{1-shot} & \textbf{2-shot} & \textbf{3-shot} & \textbf{5-shot} & \textbf{10-shot} \\ \midrule

\textbf{Real-World Baseline} & \textbf{0.3070} & 0.5445 & 0.6243 & 0.6611 & 0.7041 & 0.7667 \\ \midrule

\textbf{MDM (Synthetic)} \\
\quad $1\times$  & 0.2460 & 0.5120 & 0.6160 & 0.6559 & 0.6974 & 0.7653 \\
\quad $2\times$  & \underline{0.2928} & 0.5334 & 0.6089 & 0.6505 & 0.7008 & 0.7368 \\
\quad $4\times$  & 0.2546 & 0.5299 & 0.6344 & 0.6749 & 0.7023 & 0.7663 \\
\quad $8\times$  & 0.2654 & \textbf{0.5517} & 0.6266 & 0.6704 & 0.7165 & 0.7792 \\ \midrule

\textbf{T2M (Synthetic)} \\
\quad $1\times$  & 0.2606 & 0.5229 & 0.6100 & 0.6551 & 0.7024 & 0.7649 \\
\quad $2\times$  & 0.2462 & 0.5132 & 0.6062 & 0.6669 & \textbf{0.7226} & 0.7635 \\
\quad $4\times$  & 0.2482 & 0.5256 & 0.6114 & \underline{0.6766} & 0.7024 & 0.7719 \\
\quad $8\times$  & 0.2123 & \underline{0.5491} & \underline{0.6372} & 0.6631 & 0.7050 & 0.7891 \\ \midrule

\textbf{MotionDiffuse (Synthetic)} \\
\quad $1\times$  & 0.2738 & 0.5363 & 0.6060 & 0.6548 & 0.7053 & 0.7739 \\
\quad $2\times$  & 0.2739 & 0.5233 & 0.6186 & 0.6725 & 0.6995 & 0.7907 \\
\quad $4\times$  & 0.2566 & 0.5423 & 0.6472 & 0.6700 & 0.7150 & \textbf{0.7959} \\
\quad $8\times$  & 0.2427 & 0.5460 & \textbf{0.6550} & \textbf{0.6893} & \underline{0.7213} & \underline{0.7956} \\ 
\bottomrule
\end{tabular}
\end{table}

We examine the effect of scaling the amount of synthetic data used during pretraining. For each text-to-motion model, we pretrain using synthetic datasets at $1\times$, $2 \times$, $4 \times$, and $8 \times$ the size of the real-world dataset, and report downstream evaluation results in Table~\ref{tab:synthetic_scaling_results}.

While synthetic pretraining benefits from scale, a consistent gap remains in the 0-shot setting: even at $8\times$ scale, synthetic-pretrained models do not surpass the real-world baseline without fine-tuning. We analyse how 0-shot evaluation behaves more extensively in Section~\ref{subsubsection:ScalingRealWorldData}. However, this result seems to suggest that real-world data induces representations that generalise more effectively in the absence of task-specific supervision, whereas synthetic pretraining derives greater benefit from fine-tuning when labelled data is available.

While performance generally improves as synthetic pretraining data is scaled, the trend is not strictly monotonic. This contrasts with prior work showing monotonic gains when scaling real-world pretraining data \cite{yuan2024self}. The clear cause for this variability in the synthetic setting is unclear to us, but it may stem from stochasticity in synthetic data generation, such as differences in the semantic expressiveness of sampled text prompts or randomness in the sampled motion lengths.

\subsubsection{Scaling Real-World Data}
\label{subsubsection:ScalingRealWorldData}

\begin{figure}[t]
  \centering
  \includegraphics[
    width=0.8\linewidth,
  ]{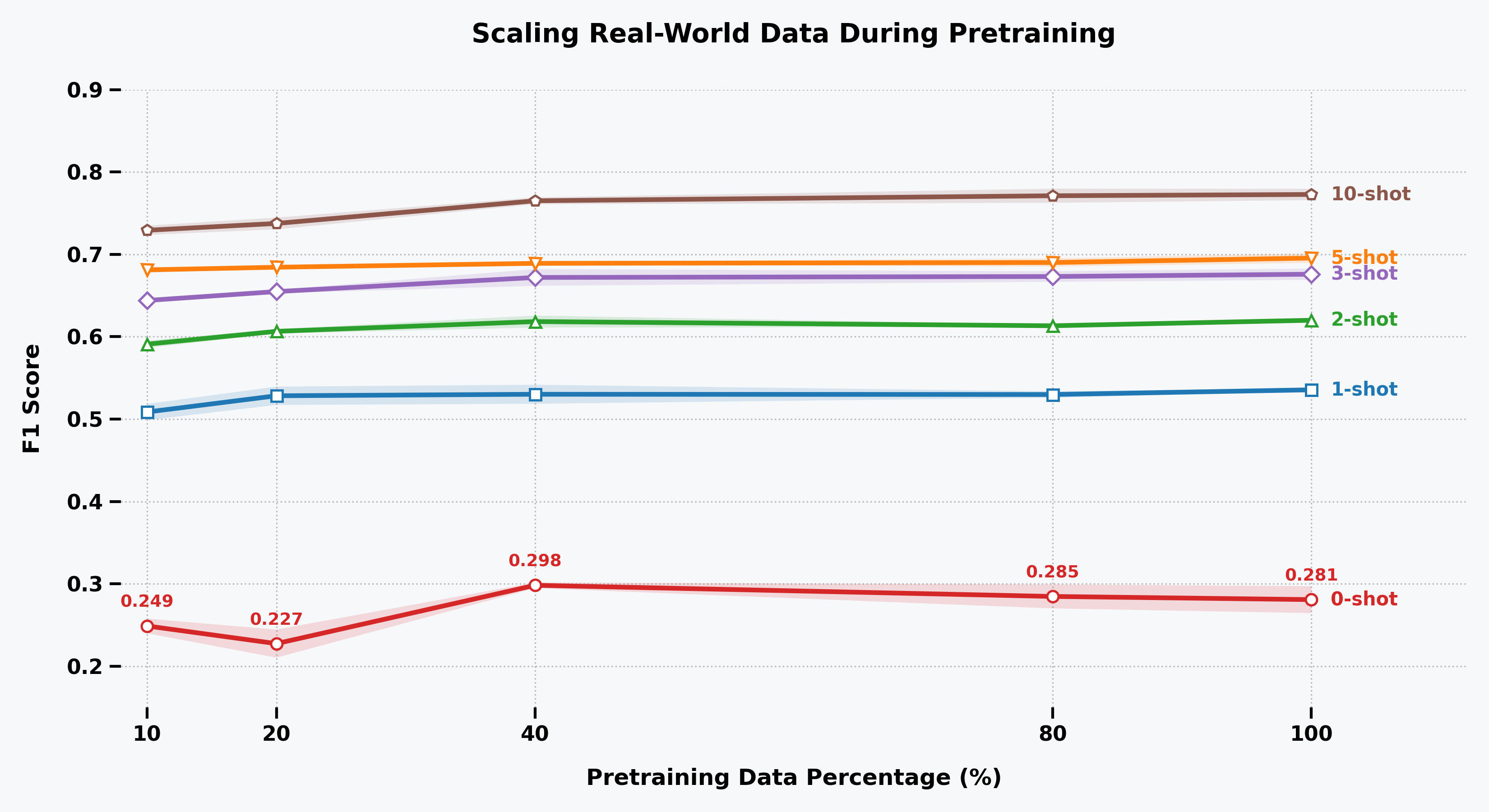}
  \caption{\textbf{Fine-tuning stabilises gains from real-world pretraining.} While increasing real-world pretraining data yields small but consistent improvements in fine-tuned ($k$-shot) performance, zero-shot performance remains unstable and does not scale reliably with data volume. Curves show results from three pretrained models per data scale, using the same set of random seeds at each percentage; shaded regions indicate standard error. Scores are mean macro-averaged F1 across downstream datasets.}
  \label{fig:ScalingRealWorldData}
\end{figure}

We study the effect of scaling real-world pretraining data by randomly subsampling 10\%, 20\%, 40\%, and 80\% of the available text–motion pairs and pretraining models on each subset, with three random seeds per configuration. This analysis serves two purposes: first, to examine the stability of $k$-shot evaluations by quantifying variability across runs, motivated by the persistent 0-shot performance gap between real and synthetic pretraining; and second, to assess whether the commonly held expectation that increased pretraining data volume leads to improved downstream performance \citep{narayanswamy2024scalingwearablefoundationmodels} holds when scaling real-world text–motion data in our setting, given that this trend did not hold for synthetic data in Section~\ref{subsubsection:PretrainingwithSyntheticData}. Results are shown in Figure~\ref{fig:ScalingRealWorldData}.

Given the high performance variability across trained models and inconsistent scaling results in the 0-shot setting, out-of-the-box generalisation may not be the best metric to fully capture the quality of learned representations for HAR tasks. It seems comparisons between pretrained motion time-series models are more reliable when some degree of fine-tuning is undertaken.

More critically, even in the fine-tuning setting, scaling real-world data yields only marginal gains, with pretraining on 40\% of the data nearly matching performance obtained using the full dataset. This stands in contrast to \cite{yuan2024self, narayanswamy2024scalingwearablefoundationmodels}, which reports substantially larger improvements when scaling real-world pretraining data, but does so using wearable sensor data rather than motion capture. Taken together, these results suggest a domain mismatch: pretraining on motion capture may emphasise factors that are weakly aligned with the signal characteristics required for wearable-based HAR, indicating that motion capture may not be the most appropriate representation for learning transferable motion representations (Figure~\ref{fig:ProblemFormulationImage}).

\section{Conclusion}
\label{ref:Conclusion}

\begin{figure}[t]
  \centering
  \includegraphics[
    width=.8\linewidth,
  ]{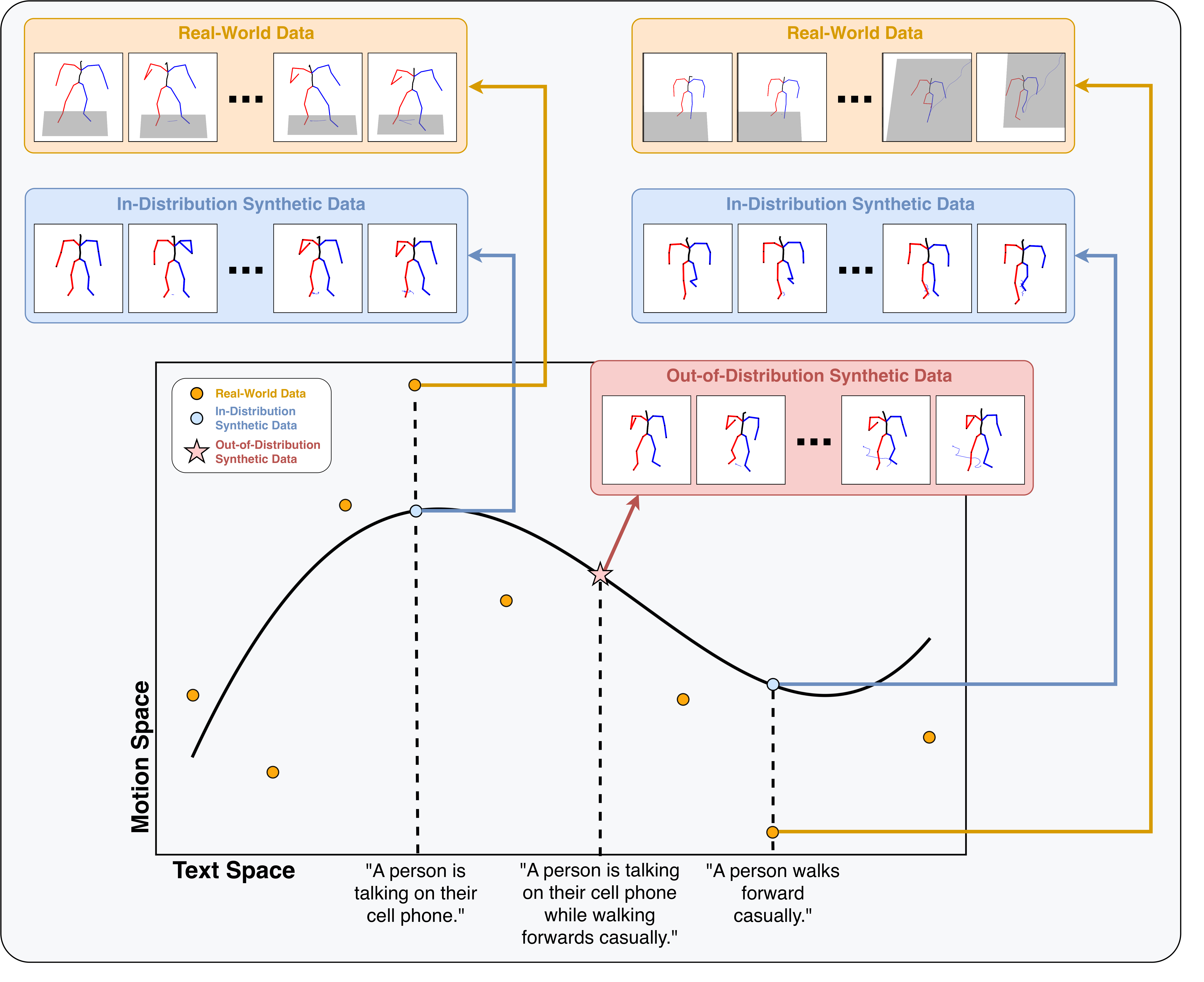}
  \caption{\textbf{Using the Text-to-Motion Model.} Assuming text and motion samples are embedded and projected into a shared low-dimensional space for visualisation, where the black curve represents the text-to-motion model (shown as deterministic mapping here). In-distribution synthetic data is obtained by sampling the text-to-motion model using matching text inputs from our real-world data. Generating out-of-distribution synthetic data consists of sampling motion from text unseen in our real-world data.}
  \label{fig:T2MasFunction}
\end{figure}

In this paper, we investigate the impact of incorporating synthetic motion data into the pretraining of motion time-series models for HAR. Our results indicate that generalisation to downstream tasks can be improved either by mixing real and synthetic data during pretraining or by using purely synthetic data at sufficiently large scales.

Our reliance on motion capture as a proxy for human motion during pretraining is a key bottleneck in motion pretraining (Figure~\ref{fig:ProblemFormulationImage}). Motion capture may not constitute an optimal surrogate or representation for human motion when the downstream objective is HAR, which may in turn constrain the benefits of synthetic data designed to approximate the motion capture domain. As a result, the domain over which our representations are learned during pretraining may be misaligned with the domain used downstream.

Looking ahead, synthetic data generation offers the potential for even larger gains if it is not restricted to the real-world text descriptions used to train existing text-to-motion models. The strength of text-to-motion models lies in their ability to interpolate over a learned motion manifold (Figure~\ref{fig:T2MasFunction}), enabling the synthesis of diverse and novel motion patterns that are absent from motion capture datasets but potentially beneficial for HAR. Exploring richer and more targeted prompt distributions represents a promising direction for future work.

\newpage


\subsubsection*{Acknowledgments}
We would like to acknowledge the Oxford Biomedical Research Computing (BMRC) facility, which we used to run the various experiments needed for this paper. This research was funded by the Rhodes Trust, and the Engineering and Physical Sciences Research Council (EPSRC). AD is supported by a range of grants from the Wellcome Trust [223100/Z/21/Z, 227093/Z/23/Z], Swiss Re, GSK, Boehringer Ingelheim, Google, National Institutes of Health’s Oxford Cambridge Scholars Program, EPSRC Centre for Doctoral Training in Health Data Science (EP/S02428X/1),  British Heart Foundation Centre of Research Excellence (grant number RE/18/3/34214), Cancer Research UK, and funding administered by the Danish National Research Foundation in support of the Pioneer Centre for SMARTbiomed. For the purpose of open access, the author(s) has applied a Creative Commons Attribution (CC BY) licence to any Author Accepted Manuscript version arising.

\bibliography{firas2026_conference}
\bibliographystyle{firas2026_conference}

\appendix
\section{Appendix}
\label{sec:Appendix}

\begin{table}[H]
\centering
\caption{\textbf{Synthetic Data Volume.} Synthetic data volume scaling relative to the real-world dataset. The multipliers indicate the ratio of generated motion sequences compared to the original motion capture set size.}
\label{tab:data_volume}
\addtolength{\tabcolsep}{6pt} 
\begin{tabular}{cc}
\toprule
\textbf{Volume Ratio} & \textbf{No. of Generated Sequences} \\ 
\midrule
$1\times$ & $24,660$  \\
$2\times$ & $49,320$  \\
$4\times$ & $98,640$  \\
$8\times$ & $197,280$ \\
\bottomrule
\end{tabular}
\end{table}



\end{document}